\documentclass[conference]{IEEEtran}
\IEEEoverridecommandlockouts
\usepackage{cite}
\usepackage{footnote}
\usepackage{amsmath,amssymb,amsfonts,algorithmic}
\usepackage{graphicx}
\usepackage{booktabs,textcomp}
\usepackage{multicol,multirow,makecell}
\usepackage{soul,color,colortbl} 
\usepackage[dvipsnames]{xcolor}
\def\BibTeX{{\rm B\kern-.05em{\sc i\kern-.025em b}\kern-.08em
    T\kern-.1667em\lower.7ex\hbox{E}\kern-.125emX}}
    
\newcommand{\modelname}{AdaTosk}
\newcommand\tb[1]{\textbf{#1}}
\newcommand\cc[1]{\cellcolor{gray!17}{#1}}
\newcommand\tc[1]{\textcolor{gray}{#1}}
\makeatletter
\renewcommand{\footnoterule}{
  \kern -3pt
  \hbox to \columnwidth{\rule{4cm}{0.4pt}\hfil}
  \kern 2.6pt
}
\makeatother

\begin{document}

\title{Soften the Mask: Adaptive Temporal Soft Mask for Efficient Dynamic Facial Expression Recognition\\ \thanks{$^\dag$ Corresponding author}}

\author{\IEEEauthorblockN{1\textsuperscript{st} Meng-zhu Li}
\IEEEauthorblockA{
\textit{Beijing Union University} \\
Beijing, China \\
limengzhu1123@gmail.com}
\and
\IEEEauthorblockN{2\textsuperscript{nd} Quanxing Zha$^\dag$}
\IEEEauthorblockA{\textit{Huaqiao University} \\
Xiamen, China \\
quanxing.zha@gmail.com}
\and
\IEEEauthorblockN{3\textsuperscript{rd} Hongjun Wu}
\IEEEauthorblockA{
\textit{Beijing University of Posts and Telecommunication} \\
Beijing, China\\
hongjun.wu@bupt.edu.cn}
}

\maketitle

\begin{abstract} %100-150 words, 142 words now
Dynamic Facial Expression Recognition (DFER) facilitates the understanding of psychological intentions through non-verbal communication. Existing methods struggle to manage irrelevant information, such as background noise and redundant semantics, which impacts both efficiency and effectiveness. In this work, we propose a novel supervised temporal soft masked autoencoder network for DFER, namely \modelname, which integrates a parallel supervised classification branch with the self-supervised reconstruction branch. The self-supervised reconstruction branch applies random binary hard mask to generate diverse training samples, encouraging meaningful feature representations in visible tokens. Meanwhile the classification branch employs an adaptive temporal soft mask to flexibly mask visible tokens based on their temporal significance. Its two key components, respectively of, class-agnostic and class-semantic soft masks, serve to enhance critical expression moments and reduce semantic redundancy over time. Extensive experiments conducted on widely-used benchmarks demonstrate that our \modelname{} remarkably reduces computational costs compared with current state-of-the-art methods while still maintaining competitive performance.
\end{abstract}

% \vspace{-0.5em}
\section{Introduction}
\label{sec:intro}

Facial expression recognition (FER), which aims to accurately identify facial expressions from static visual contents, has emerged as a vital tool for understanding and conveying psychological intentions, offering profound insights into human behavior and decision-making. Nevertheless, FER struggles to effectively address intractable dynamic expression variations. It is acknowledged that dynamic facial expression recognition (DFER)~\cite{dfer,guo2024smile,xiao2024estme} is therefore a promising direction as it well accounts for spatiotemporal signals in videos, potentially enabling dynamical human expressions in real-world scenarios. 

However, the efficiency of DFER still remains significant challenge due to substantial redundancy inadvertently introduced by temporal continuity in videos. Inspired by the remarkable success of Masked AutoEncoder (MAE)~\cite{he2022masked}, some attempts~\cite{liang2022supmae,li2023dr} have improved the efficiency by
utilizing the visible patches that have potential to reconstruct the original content. Building on this foundation, recent efforts~\cite{sun2023maedfer,sun2024svfap} further reduce the computational costs during the reconstruction process with informative features learned by efficient encoders, significantly advancing the efficiency and efficacy of DFER. Despite these advancements, a critical limitation remains: the tendency to omit subtle and hard-to-detect expression dynamics in videos.

\begin{figure}[t]
\centerline{\includegraphics[scale=0.65]{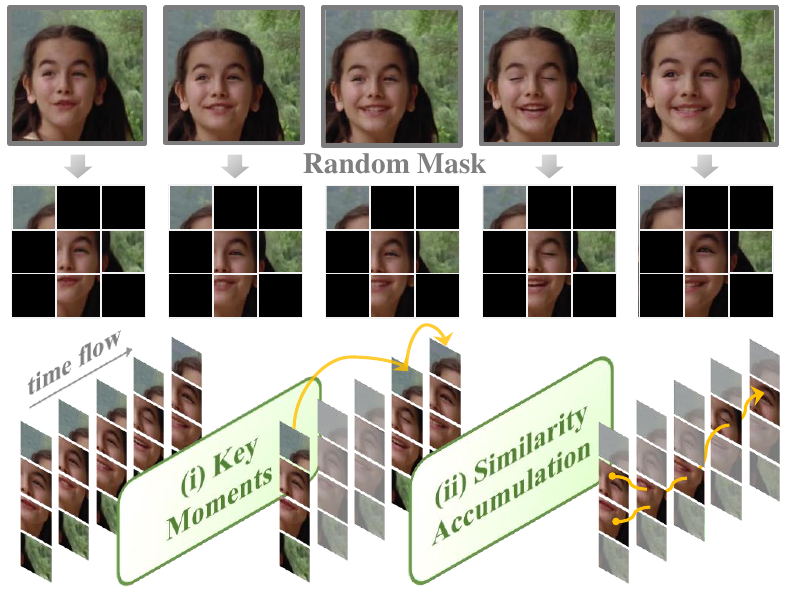}}
\caption{\textbf{Redundancy in visible tokens.} Row 1: All original frames. Row 2:\\ Frames after random masking. Row 3: \colorbox{gray!17}{Gray-masked} regions indicate redundant temporal information in visible tokens. Yellow lines connect all distinct frames/tokens. (i) Key moments represent the important frames; (ii) Similarity accumulation shows resemble tokens exist redundancy in neighbor time.}
\vspace{-2em}
\label{fig:teaser}
\end{figure}

Different from images, video data exhibits not only spatial redundancy but also additional temporal redundancy. Although the reconstruction branch removes many redundant tokens, the remaining visible tokens still contain much irrelevant information, such as background noise.  As shown in Fig.~\ref{fig:teaser} Row 3, expression dynamics in the DFER task are sparse and typically triggered by a few salient frames. And these dynamics are generally less pronounced than those in action recognition, occurring mainly in key facial regions, such as the mouth. Hence, there is considerable potential for improving efficiency in the DFER task. Our main goals are: (i) To focus on key moments of expression dynamics while avoiding unnecessary computation on irrelevant frames;
(ii) To reduce the temporal accumulation of semantically similar tokens, such as non-essential actions and static information.

To alleviate temporal redundancy, the intuitive ideal is amplifying the influence of temporally valuable tokens while reducing the impact of meaningless ones. Inspired by the binary masking mechanism in MAE~\cite{he2022masked}, we design an adaptive masking strategy that applies varying degrees of masking to the visible tokens. For example, we employ lighter masks to informative tokens, preserving their influence. To distinguish these two types of masking, we refer to the binary mask as \textbf{`Hard Mask'} and the adaptive mask  as \textbf{`Soft Mask'}. Therefore, a novel adaptive temporal soft mask mechanism is introduced for aforementioned goals. (i) To preserve the representation of expression dynamics without being constrained by predefined categories, we design a \textit{class-agnostic dynamic soft mask}. By calculating frame-to-frame feature differences, we select and enhance activated frames, with the soft mask defined by their influence on the entire time sequence. (ii) Some tokens may be temporally repetitive but still carry essential information of expression. To address this, we propose a \textit{class-semantic similar soft mask} to integrate class semantics into similarity accumulation over time. To prevent the soft mask from overemphasizing specific tokens, we exclude some high-score tokens from the last frame during transitions.

In light of these, we propose a novel \textbf{Ada}ptive \textbf{T}emporal S\textbf{o}ft Ma\textbf{sk} mechanism, namely \textbf{AdaTosk}, for Efficient Dynamic Expression Recognition. The overview of our \modelname{} is illustrated in Fig.~\ref{fig:model} and it mainly contains two parallel branches. Specifically, the self-supervised reconstruction branch applies random hard mask for data augmentation, generating diverse training samples and promoting meaningful representations in visible tokens in each iteration. Corresponding to the hard mask, the parallel supervised classification branch employs a unique soft mask mechanism, assigning adaptive masks to visible tokens based on their temporal redundancy and semantics in the video. The two core strategies, class-agnostic and class-semantic soft masks, respectively capture key moments in expression dynamics and the accumulation of semantic redundancy over time.

\textbf{Our contributions can be summarized as following:}  (i) Our \modelname{} is the first to integrate supervised VideoMAE into the DFER task, where the optimization of visible tokens significantly enhances the efficiency of expression recognition; (ii) We propose an innovative adaptive temporal soft mask that adaptively adjusts the masking degree. Its two designs: the class-agnostic dynamic and class-semantic similar soft masks, respectively activate key expression dynamics and suppress semantic redundancy over time; (iii) \modelname{} achieves a 3\% performance improvement over state-of-the-art methods while reducing the model size by about 10M parameters and lowering computational cost by 6G FLOPs on the FERV39K~\cite{wang2022ferv39k}, DFEW~\cite{jiang2020dfew}, and MAFW~\cite{liu2022mafw} datasets.

\section{Related Work}
\label{sec:releated}
\textbf{Dynamic facial expression recognition.} Early deep learning models~\cite{jiang2020dfew,wang2022ferv39k,m3dfel2023,ial2023,zha2024ugncl} introduce end-to-end supervised architectures capable of extracting distinct spatial and temporal features. Later, Transformer-based models~\cite{liu2022mafw,former-dfer2022}, leveraging ViT~\cite{dosovitskiy2020image} encoders, have become dominant in this field. Recently, there has been a notable trend toward large-scale models pre-trained on massive datasets in a self-supervised manner. Multi-modal vision-language models~\cite{cliper2024,dfer-clip2024,emoclip2024,a3} have become essential for the DFER task. Additionally, self-supervised pre-training methods~\cite{sun2024svfap,sun2023maedfer} utilizing VideoMAE~\cite{tong2022videomae} with large-scale unlabeled datasets focus on masked reconstruction pre-training for facial video data. Building on this foundation, approaches~\cite{sun2024hicmae,mma-dfer2024} further adapt these models to support multi-modal inputs. Nevertheless, these models rely heavily on detailed annotations or substantial computation during fine-tuning. Inspired by SupMAE~\cite{liang2022supmae}, our model incorporates a parallel supervised branch that performs classification using only visible tokens, significantly improving efficiency. Meanwhile, DR-FER~\cite{li2023dr}, which is based on SupMAE~\cite{liang2022supmae}, further validates the effectiveness of this framework in static FER.

\textbf{Video Spatio-Temporal Redundancy.} Reducing spatio-temporal redundancy for efficient video analysis can be approached through three main strategies. (i) Adaptive frame selection~\cite{2d2021} dynamically identifies the most relevant frames for recognition tasks, while (ii) spatio-temporal region localization models~\cite{adaptive2021,adafocus2022} focus on identifying the most task-relevant regions within videos. Additionally, (iii) token pruning methods in Transformers, such as DynamicViT~\cite{dynamicvit2021} and EViT~\cite{evit2022}, eliminate less significant tokens using class token attention, and ToMe~\cite{merge2022} accelerates computation by merging similar tokens.  Inspired by these approaches, we propose a adaptive temporal soft mask mechanism to mitigate the temporal redundancy of expression features in videos. By applying varying degrees of masking based on token influence, this approach not only prevents information loss but also reduces redundancy.  

\section{Methodology}
\label{sec:method}
\begin{figure*}[t]
\centerline{\includegraphics[scale=0.61]{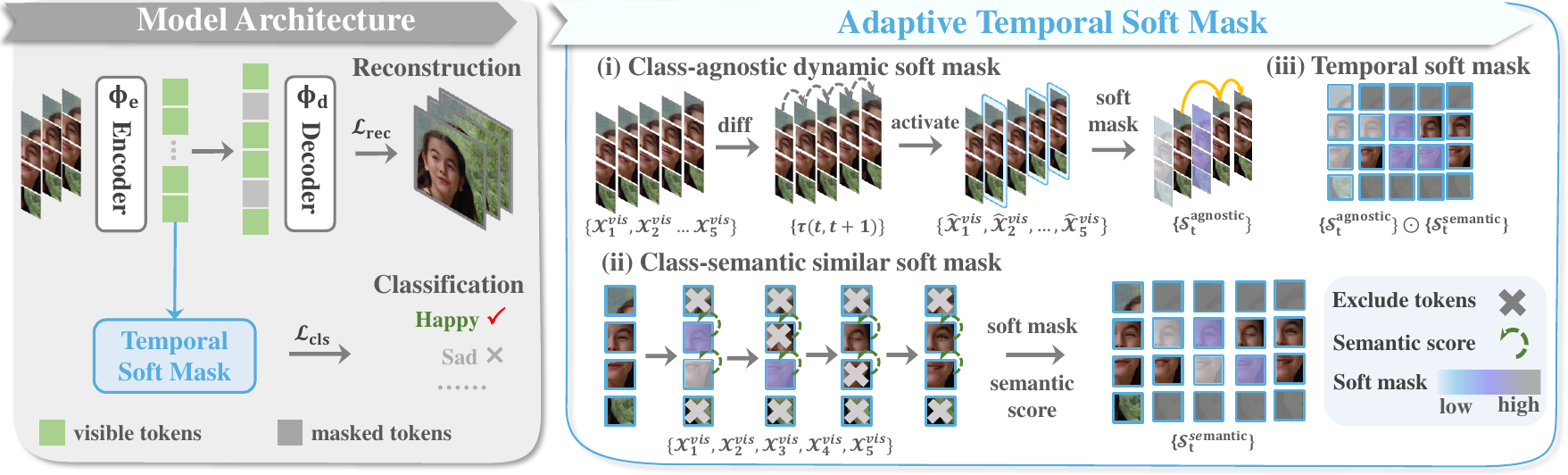}}
\vspace{-0.5em}
\caption{\textbf{Overall framework of \modelname.} The \colorbox{gray!17}{left part} illustrates the overall model architecture with two parallel branches. After the hard mask, the \textcolor{OliveGreen}{visible tokens} are processed by the encoder. Then the decoder reconstructs the masked tokens using the reconstruction loss $\mathcal{L}_{rec}$, meanwhile a temporal soft mask is applied for expression recognition with the classification loss $\mathcal{L}_{cls}$. The \colorbox{cyan!8}{right part} represents the adaptive temporal soft mask mechanism: (i) Class-agnostic dynamic soft mask identifies key frames; (ii) Class-semantic similar soft mask transfers the accumulative temporal score between adjacent frames; (iii) The final temporal soft mask combines results from (i) and (ii).}
\vspace{-1.5em}
\label{fig:model}
\end{figure*}

\textbf{Overview.} The architecture of \modelname{} is illustrated in Fig.~\ref{fig:model}. Our model achieves efficient facial expression recognition by focusing on the most informative visible tokens. Given a video token set $\mathcal{X}$, the goal is to extract the optimal visible tokens ${\mathcal{X}}^\text{opt}$ by integrating a random hard mask $\mathcal{H}$ (self-supervised branch) and a temporal soft mask $\mathcal{S}$ (supervised branch). For the $s$-th token in the $t$-th frame, $\mathcal{X}_{t,s}$ is element-wise multiplied by $\mathcal{H}_{t,s}$ and $\mathcal{S}_{t,s}$, defined as:
\begin{equation}\label{eq:1}
\begin{aligned}
&\mathcal{X}^\text{vis}_{t,s} = \mathcal{X}_{t,s} \odot \mathcal{H}_{t,s},
&\mathcal{X}^\text{opt}_{t,s} = \mathcal{X}^\text{vis}_{t,s} \odot \mathcal{S}_{t,s}.
\end{aligned}
\end{equation}
\noindent The self-supervised branch applies reconstruction loss $\mathcal{L}_\text{rec}$ to measure the distance between the reconstructed result $\widetilde{\mathcal{X}}$ and original token set $\mathcal{X}$. The supervised branch classifies the expression utilizes optimal token set $\mathcal{X}^\text{opt}$ with classification loss $\mathcal{L}_\text{cls}$. The total loss $\mathcal{L}$ is therefore defined as:
\begin{equation}\label{eq:2}
  \mathcal{L} = \lambda_\text{rec} \mathcal{L}_\text{rec}(\mathcal{X}, \widetilde{\mathcal{X}}) + \lambda_\text{cls} \mathcal{L}_\text{cls}({\mathcal{X}}^\text{opt}),
\end{equation} 
\noindent where $\lambda_\text{rec}$ and $\lambda_\text{cls}$ are balanced weights and set as 1 and 0.1.
\subsection{Random Binary Hard Mask}
Given a video $\mathcal{V} \in \mathbb{R}^{T \times H \times W \times C}$, where $T$, $H$, $W$, and $C$ denote the number of frames, height, width, and RGB channels, the video is first divided into patches of size $h \times w \times C$. These patches are then grouped across $t$ consecutive frames to form tokens of size $t \times h \times w \times C$. After embedding, the video is transformed into tokens set $\mathcal{X} \in \mathbb{R}^{N_t \times N_s \times D}$, where $N_t = \frac{T}{t}$ is the number of temporal tokens, $N_s = \frac{H \times W}{h \times w}$ is the number of spatial patches per frame, and $D$ is the token dimension derived from $t \times h \times w \times C$.

The random hard mask mechanism is derived from VideoMAE~\cite{tong2022videomae} which can reconstruct masked tokens from the visible ones. This self-supervised strategy encourages the model to focus on visible tokens containing critical features. The binary hard mask $\mathcal{H}_t \in \{0, 1\}^{N_s \times 1}$ is generated spatially and then broadcast across the temporal axis. The value of 1 remains tokens as visible $\mathcal{X}^\text{vis}$, while 0 masks them as $\mathcal{X}^\text{mask}$. As shown in \eqref{eq:3}, the high-capacity encoder $\Phi_e$ processes the visible tokens, while the lightweight decoder $\Phi_d$ reconstructs the masked tokens using encoded visible tokens and learnable representations. The reconstruction loss $\mathcal{L}_\text{rec}$ is defined as the Mean Squared Error (MSE) between the original tokens $\mathcal{X}$ and the reconstructed tokens, with function $\Psi$ identifying masked positions in the pixel space.
\begin{equation} \label{eq:3}
\mathcal{L}_\text{rec} = \textrm{MSE}(\Phi_d(\Phi_e(\mathcal{X}^\text{vis}) \cup \mathcal{X}^\text{mask}), \mathcal{X} \odot \Psi(1-\mathcal{H}))
\end{equation}

\subsection{Adaptive Temporal Soft Mask}
To assess the contribution of tokens to expression evolution over time, we introduce an adaptive temporal soft mask. Tokens that significantly influence expression changes receive a lighter mask, while less relevant tokens are assigned a heavier mask. This adaptive masking is guided by two factors: (i) the extent of expression change throughout the temporal sequence, and (ii) the accumulation of semantic similarity among tokens over time. Based on these principles, we propose two soft mask strategies. The temporal sequence length is $n_t$=$N_t$, and the number of visible tokens is $n_s$=$N_s\times$$(1-\rho)$, where $\rho$ is hard mask ratio.

\textbf{Class-agnostic dynamic soft mask.} Since expression changes occur gradually, they can be captured through several key moments. To avoid biasing expression dynamics by individual differences or predefined categories, we use a class-agnostic soft mask to analyze feature variations between neighboring frames. The temporal feature difference $\tau(t-1, t)$ for each adjacent frame pair $\{\mathcal{X}^\text{vis}_{t-1}, \mathcal{X}^\text{vis}_t\}$ is calculated as:
\begin{equation}\label{eq:4}
    \tau(t-1, t) = \sum_{d=1}^{D}| \text{diff}(\mathcal{X}^\text{vis}_{t-1}, \mathcal{X}^\text{vis}_t, d) |,
\end{equation}
\noindent where $\mathcal{X}^\text{vis}_t = \text{avgpool}(\sum_{s=1}^{n_s}\mathcal{X}^\text{vis}_{t,s})$.  `$\text{diff}(\cdot)$' represents element-wise subtraction, and $d$ is the feature index. We then select the top-K frame pairs from set $\tau$ and mark the second frame in each pair as the activate frame $\widehat{\mathcal{X}}^\text{vis}_t$.
\begin{equation}\label{eq:5}
\tau = \{ \tau(1,2), \tau(2,3),\dots, \tau(t-1,t)\}; \; \text{TopK}(\tau)=1.
\end{equation}
In order to increase the representation of activate frames, we leverage a channel-wise information interaction unit in the squeeze-and-excitation pattern to generate the feature $\mathcal{\widetilde{X}}^\text{vis}_t$.
\begin{equation}\label{eq:6}
\widetilde{\mathcal{X}}^\text{vis}_t = \frac{\exp(\theta(\widehat{\mathcal{X}}^\text{vis}_t)) }{\sum_{d=1}^{D} \exp(\theta(\widehat{\mathcal{X}}^\text{vis}_t) )}\otimes \widehat{\mathcal{X}}^\text{vis}_t +\widehat{\mathcal{X}}^\text{vis}_t,
\end{equation}
where $\otimes$ denotes element-wise multiplication. The function $\theta$ is a simple multi-layer perceptron (MLP) with two fully connected (FC) layers and a ReLU activation in between. The first FC layer has a weight matrix $W_1 \in \mathbb{R}^{D \times (D / \mu)}$, while the second uses $W_2 \in \mathbb{R}^{(D / \mu) \times D}$, where $\mu$ is a scaling factor. A residual connection is added to improve training stability. A temporal information interaction unit is applied to capture global contextual relationships between $\mathcal{\tilde{X}}^\text{vis}_t$ and $\mathcal{X}^\text{vis}$, as defined below:
\begin{equation}\label{eq:7}
  \mathcal{S}^\text{agnostic}_t = \text{softmax}(\mathcal{X}^\text{vis} \odot (\widetilde{\mathcal{X}}^\text{vis}_t)^T), 
\end{equation}
where $\mathcal{S}^\text{agnostic}_t \in \mathbb{R}^{n_s \times D}$ is our class-agnostic soft mask.

\textbf{Class-semantic similar soft mask.} Although the hard mask mechanism eliminates many tokens, redundant information remains in the visible tokens, such as the static environmental tokens that persist across neighboring frames. Therefore, we propose a class-semantic similar soft mask, calculating token similarities frame-by-frame. We define the temporal accumulative score $\mathcal{M} \in [0,1]^{n_t \times n_s}$ to model the probability of masking a token in the $t$-th frame.
\begin{equation}\label{eq:8}
\mathcal{M}_{t,s} = \mathbb{P}_{\text{mask}}(\mathcal{X}^\text{vis}_{t,s}) \in [0, 1] \quad \text{s.t.} \quad \sum_{s=1}^{n_s} \mathcal{M}_{t,s} = 1.
\end{equation}
As time progresses, certain tokens accumulate significant higher scores than others. To reduce computational cost and better detect global similarities, we exclude the top $r$ tokens from $\mathcal{X}^\text{vis}_t$ when calculating $\mathcal{M}_{t+1}$, resulting $\mathcal{X'}^\text{vis}_t \in \mathbb{R}^{(n_s-r) \times D}$. 
Considering continuity in redundancy analysis across frames, we apply a Markov chain for transferring the accumulative score $\mathcal{M}'_t \in \mathbb{R}^{(n_s - r) \times 1}$ to $\mathcal{M}_{t+1}$ via the transition probability $\mathbb{P}_\text{mask}(\mathcal{X}^\text{vis}_t | \mathcal{X'}^\text{vis}_{t-1}) \in \mathbb{R}^{n_s \times (n_s - r)}$.
 Mathematically, 
\begin{equation}\label{eq:9}
\begin{aligned}
&\mathcal{M}_{t+1} =  \mathbb{P}_\text{mask}(\mathcal{X}^\text{vis}_{t+1}|\mathcal{X'}^\text{vis}_t)\mathcal{M}'_t,\\
&\mathbb{P}_{\text{mask}}(\mathcal{X}^\text{vis}_{t+1}|\mathcal{X'}^\text{vis}_t)= \text{softmax}(f(\mathcal{X}^\text{vis}_{t+1}) f(\mathcal{X'}^\text{vis}_t)^T ),
\end{aligned}
\end{equation}
where $f$ is the projection head for similarity, and the transition probability is derived from a softmax affinity matrix. Although the Markov chain has reduce much redundancy, some similar tokens may contain crucial semantic information that improves model performance. We define the activated attention map $\mathcal{A}$ to compute the semantic score for token $\mathcal{X}^\text{vis}_\text{t,s}$ as:
\begin{equation}\label{eq:10}
\mathcal{A}(\mathcal{X}^\text{vis}_{t,s})= \sum_{d=1}^{D} |\mathcal{X}^\text{vis}_{t,s,d}| \in \mathbb{R}^+.
\end{equation}
Under the class supervision, we sum absolute activations across channels to highlight contributions to subsequent layers, using these activation-based maps (Fig.~\ref{fig:vis}) to adjust temporal scores. This re-weights the temporal accumulative score $\mathcal{M}$, ensuring that tokens with high semantic value are prioritized for retention, even if they exhibit temporal redundancy.

Finally, the class-semantic similar soft mask $\mathcal{S}^\text{semantic}_{t,s}$ is generated from the temporal accumulative score $\mathcal{M}_{t,s}$ and the semantic score  $\mathcal{A}(\mathcal{X}^\text{vis}_{t,s})$.
\begin{equation}\label{eq:11}
\mathcal{S}^\text{semantic}_{t,s} =  \mathcal{A}(\mathcal{X}^\text{vis}_{t,s})(1-\mathcal{M}_{t,s}),
\end{equation}
where $\mathcal{S}^\text{semantic}_{t,s}$ is min-max normalized to the range [0,1]. And \textbf{The final adaptive temporal soft mask $\mathcal{S}$} is defined as:
\begin{equation}\label{eq:12} 
\mathcal{S}_{t,s} = \mathcal{S}^\text{agnostic}_t \odot \mathcal{S}^\text{semantic}_{t,s}. 
\end{equation} 
To classify the optimal visible tokens ${\mathcal{X}}^\text{opt}$ introduced in overview, we use the cross-entropy loss as the classification loss $\mathcal{L}_\text{cls}$:
\begin{equation}\label{eq:13} \mathcal{L}_\text{cls} = -\sum_{i=1}^{N} z_i \log(p_i), 
\end{equation}
where $p$=$ \text{avgpool(${\mathcal{X}}^\text{opt}$)}\in \mathbb{R}^N$ represents the model's prediction for $N$ classes, and $z$ is the one-hot ground truth label.

\section{Experiment}
\label{sec:exp}
\subsection{Experimental Settings}
\begin{table*}[t]
\caption{\tb{Comparison of results on three datasets.} \colorbox{gray!17}{Gray rows} indicate large language model methods, and \colorbox{cyan!8}{blue rows} represent VideoMAE-based methods. \tb{Bold values} are the best results, and \ul{underlined values} are the second best. \tc{\tb{\modelname}} and $\tb{\modelname}^\dag$ denote models with and without the temporal soft mask.}
\setlength\tabcolsep{7pt}
\centering
\begin{tabular}{lccccccccc}
\toprule
\multirow{2}{*}{\tb{Method}}  & \multirow{2}{*}{\tb{Backbone}} & \multirow{2}{*}{\tb{\makecell{\#Params\\(M)}}} & \multirow{2}{*}{\tb{\makecell{FLOPs\\(G)}}} & \multicolumn{2}{c}{\tb{DFEW}} & \multicolumn{2}{c}{\tb{FERV39K}} & \multicolumn{2}{c}{\tb{MAFW}} \\
\cmidrule(lr){5-6} \cmidrule(lr){7-8} \cmidrule(lr){9-10}
&&&& {UAR} & {WAR} & {UAR} & {WAR} & {UAR} & {WAR} \\
\midrule
\multicolumn{10}{l}{\textit{\tb{Supervised methods}}} \\
EC-STFL (MM'20)~\cite{jiang2020dfew}                                & C3D / P3D           & 78  & 8  & 45.35 & 56.51 & -     & -     & -     & -     \\
Former-DFER (MM’21)~\cite{former-dfer2022}                          & Transformer         & 18  & 9  & 53.69 & 65.70 & 37.20 & 46.85 & 31.16 & 43.27 \\
T-ESFL (MM'22)~\cite{liu2022mafw}                                   & ResNet-Transformer  & -   & -  & -     & -     & -     & -     & 33.28 & 48.18 \\
Freq-HD (MM’23)~\cite{wang2022ferv39k}                              & VGG13-LSTM          & -   & 9  & -     & -     & 32.79 & 44.54 & -     & -     \\
M3DFEL (CVPR'23)~\cite{m3dfel2023}                                  & ResNet-18-3D        & -   & 2  & 56.10 & 69.25 & 35.94 & 47.67 & -     & -     \\
% AEN (CVPRW'23)~\cite{aen2023}                                       & ResNet-18           & -   & -  & 56.66 & 69.37 & 38.18 & 47.88 & -     & -     \\
IAL (AAAI'23)~\cite{ial2023}                                        & ResNet-18           & 19  & 10 & 55.71 & 69.24 & 35.82 & 48.54 & -     & -     \\ 
\midrule
\multicolumn{10}{l}{\textit{\tb{Self-supervised methods}}} \\ 
\rowcolor{gray!17!} DFER-CLIP (BMVC'23)~\cite{dfer-clip2024}        & CLIP-ViT-B/32       & 90  & -  & 59.61 & 71.25 & 41.27 & 51.65 & 39.89 & 52.55 \\ 
\rowcolor{gray!17!} CLIPER (ICME'24)~\cite{cliper2024}              & CLIP-ViT-B/16       & 88  & -  & 57.56 & 70.84 & 41.23 & 51.34 & -     & -     \\
\rowcolor{gray!17!} EmoCLIP (FG’24)~\cite{emoclip2024}              & CLIP-ViT-B/32       & -   & -  & 58.04 & 62.12 & 31.41 & 36.18 & 34.24 & 41.46 \\
\rowcolor{gray!17!} A$^{3}$lign-DFER (ArXiv’24)~\cite{a3}           & CLIP-ViT-L/14       & -   & -  & 64.09 & 74.20 & 41.87 & 51.77 & 42.07 & 53.24 \\
\rowcolor{cyan!8!} VideoMAE (NeurIPS'22)~\cite{tong2022videomae}    & ViT-B/16            & 86  & 82 & 63.60 & 74.60 & -     & -     & 42.87 & 53.51 \\
\rowcolor{cyan!8!} MAE-DFER (MM'23)~\cite{sun2023maedfer}           & ViT-B/16            & 85  & 50 & 62.59 & 74.88 & \ul{43.12} & 52.07 & 41.62 & 54.31 \\
\rowcolor{cyan!8!} SVFAP (TAC'24)~\cite{sun2024svfap}               & ViT-B/16            & 78  & 44 & 62.63 & 74.81 & 42.14 & \ul{52.29} & 41.19 & 54.28 \\
\rowcolor{cyan!8!} HICMAE (Fusion'24)~\cite{sun2024hicmae}          & ViT-B/16            & 81  & 46 & 63.76 & 75.01 & -     & -     & 42.65 & 56.17 \\
\rowcolor{cyan!8!} MMA-DFER (CVPRW'24)~\cite{mma-dfer2024}          & ViT-B/16            & -   & -  & \tb{67.01} & \ul{77.51} & -     & -     & \ul{44.11} & \tb{58.52} \\
\rowcolor{cyan!8!} \tc{\tb{\modelname} \tb{ (Ours)}}                & \tc{ViT-B/16}       & \tc{74} & \tc{44} & \tc{64.93} & \tc{75.95} & \tc{44.34} & \tc{54.24} & \tc{42.26} & \tc{56.81} \\  
\rowcolor{cyan!8!} $\tb{\modelname}^\dag$ \tb{(Ours)}               & ViT-B/16            & 72  & 40 & \ul{66.76} & \tb{77.54} & \tb{46.28} & \tb{56.52} & \tb{44.15} & \ul{57.92} \\
\bottomrule
\end{tabular}
\vspace{-2em}
\label{tab:sota}
\end{table*}

%#################################################
% Modules ablation
%#################################################
\begin{table}[t]
\setlength\tabcolsep{17pt}
\caption{Soft mask mechanisms.}
\centering \label{subtab:module}
\begin{tabular}{lcc}
\textbf{Soft mask strategy}             & \textbf{UAR (\%)}        & \textbf{WAR (\%)}   \\
\midrule
CA                             & 65.55           & 76.64      \\ 
CS ($\mathcal{A}$+$\mathcal{M}$)                          & 66.03           & 77.11      \\ 
CS ($\mathcal{M}$)         & 65.78           & 76.67      \\ 
CS ($\mathcal{A}$)         & 65.95           & 76.93      \\ 
CA + CS ($\mathcal{M}$)    & 66.41           & 77.07      \\ 
CA + CS ($\mathcal{A}$)    & 66.62           & 77.31      \\ 
\tb{CA + CS ($\mathcal{A}$+$\mathcal{M}$)}               & \cc{66.76}      & \cc{77.54}\\ 
% \bottomrule
\end{tabular}
\vspace{-1em}
\end{table}
%#################################################
% Distance losses
%#################################################
\begin{table}[t]\setlength\tabcolsep{20pt}
\caption{Distance losses.}
\label{subtab:loss}
\centering 
\begin{tabular}{lccc}
\textbf{Distance loss}                                  & \textbf{UAR (\%)}   & \textbf{WAR (\%)}   \\
\midrule
\rule{0pt}{1.2em}classification                & 61.86      & 72.67      \\
\rule{0pt}{1.2em}cosine                        & 62.77      & 73.52      \\
\rule{0pt}{1.2em}L$_2$                         & 64.83      & 75.68      \\
\rule{0pt}{1.2em}\tb{L}$_1$\tb{(ours)}         & \cc{66.76} & \cc{77.54} \\
\end{tabular}
\vspace{-2em}
\end{table}
%#################################################
% Hard mask ratios and top numbers
%#################################################
\begin{table}[t]\setlength\tabcolsep{8pt}
\caption{Hard mask ratios and the number of excluded tokens (top-`r').}
\label{subtab:top}
\centering
\begin{tabular}{clcccc}
\textbf{Hard mask}   & \textbf{Metrics} & \textbf{r=0} & \textbf{r=2} & \textbf{r=4} & \textbf{r=6} \\ 
\midrule
\multirow{2}{*}{25\%} & WAR              & 76.74        & 76.68        & 76.57        & 76.08 \\ 
                      & GFLOPs           & 136.15       & 118.54       & 99.75        & 68.79 \\ 
\midrule
\multirow{2}{*}{50\%} & WAR              & 77.06        & 77.02        & 76.97        & 76.45 \\ 
                      & GFLOPs           & 84.32        & 72.24        & 60.05        & 42.13 \\ 
\midrule
\tb{70\%}& WAR              & 77.63        & \cc{77.54}& -& -\\ 
\tb{(ours)}       & GFLOPs           & 57.73        & \cc{40.12}& -& -\\ 
\end{tabular}
\vspace{-2em}
\end{table}

\textbf{Datasets and evaluation metrics.} We evaluate our method on three in-the-wild DFER datasets. DFEW\cite{jiang2020dfew} and FERV39k\cite{wang2022ferv39k} include 11,697 and 38,935 videos, respectively, with 7 expression classes. MAFW\cite{liu2022mafw} consists of 9,172 videos spanning 11 expression categories. UAR (Unweighted Average Recall) and WAR (Weighted Average Recall) are used as evaluation metrics.

\textbf{Implementation details.} Videos are sampled at 16 frames with a temporal stride of 4 and resized to $224 \times 224$ pixels. The frames are tokenized into 3D tokens of size $\{2 \times 16 \times 16\}$. In the self-supervised branch, we adopt the ViT-B/16 (512-dim)~\cite{dosovitskiy2020image} as encoder and one-layer transformers as decoder. In the supervised branch, a subset of visible tokens is processed with global pooling and classified by a two-layer MLP. During training, we primarily follow the hyper-parameter settings of VideoMAE~\cite{tong2022videomae}. Specifically, we set the hard mask ratio to 70\% and configure temporal soft masks with top-k=4, top-r=2. The model is optimized using AdamW with $\beta_1 = 0.9$, $\beta_2 = 0.95$, a batch size of 128, a base learning rate of $1.5 \times 10^{-4}$, and a weight decay of 0.05. The learning rate is scaled linearly as $\text{lr} = \text{base\_lr} \times \frac{\text{batch\_size}}{256}$. Models are trained for 100 epochs on 8 A100 GPUs, including 10 warm-up epochs. 

\subsection{Comparison with State-of-the-Art Methods}
\tb{Main results.} In Table~\ref{tab:sota}, we compare \modelname$^\dag$ with state-of-the-art methods on DFEW, FERV39K, and MAFW. Our model outperforms supervised models by 9–10\% in both metrics across all datasets. Compared to the best large language model (A$^{3}$lign-DFER), it achieves improvements of 2.7\% and 2.3\% in (UAR and WAR) on DFEW, 3.4\% and 4.8\% on FERV39K, 2.1\% and 4.7\% on MAFW. Compared to VideoMAE-based models (MAE-DFER and MMA-DFER), \modelname$^\dag$ achieves improvements of 3-4\% over the MAE-DFER across all datasets, and 3.1\% and 4.4\% over the MMA-DFER on FERV39K. Detailed results for each expression are provided in Appendix Tables 1–3.

\tb{Model efficiency.} (i) Compared to the baseline VideoMAE \cite{tong2022videomae}, our \tc{\tb{\modelname}} and \modelname$^\dag$ replace the fine-tuning stage with the parallel visible tokens classification. Although \tc{\tb{\modelname}} is without the temporal soft mask, it reduces parameters by 12M and FLOPs by 38G, demonstrating the efficiency of the architecture based on SupMAE\cite{liang2022supmae}. (ii) Compared to \tc{\tb{\modelname}}, \modelname$^\dag$ incorporates a temporal soft mask, further reducing parameters by 2M and FLOPs by 4G, while improving UAR and WAR by about 2\% across three datasets, confirming the effectiveness of the temporal soft mask. (iii) Both \tc{\tb{\modelname}} and \modelname$^\dag$ reduce parameters by 4-12M and FLOPs by 2-38G compared to other ViT-B models.

\subsection{Ablation Study}
We conduct the ablations on DFEW dataset.
\tb{Soft mask mechanism.} In Table~\ref{subtab:module}, we evaluate the effect of soft masks using several variants: (i) Rows 1-2: class-agnostic (CA) soft mask or class-semantic (CS) soft mask with two scores; (ii) Rows 3-4: redundant score $\mathcal{M}$ or semantic score $\mathcal{A}$ from the class-semantic soft mask; (iii) Rows 5-6: class-agnostic soft mask combined with one score; (iv) Row 7: CA soft mask and CS soft mask with two scores. The results show that models with both soft masks outperform the CA soft mask by 2\% and the CS soft mask by 1\%. Using a single score instead of two in the CS soft mask reduces performance by 0.1\%-0.3\%. 

\tb{Distance loss.} In Table~\ref{subtab:loss}, we compare strategies for calculating the difference between neighboring features in the class-agnostic soft mask. The `classification' strategy uses a pre-trained head to categorize frames based on distinctiveness. The results show that L$_2$ distance outperforms cosine distance, with L$_1$ distance yielding the best overall performance. We adopt L$_1$ as the default \texttt{diff} function.

\tb{Hard mask ratios and the number of excluded tokens.} In  Table~\ref{subtab:top}, we firstly analyze the impact of excluding the top-score $r$ tokens per frame when calculating token similarity in the class-semantic soft mask. At the 70\% hard mask ratio, setting $r$=2 (r=4 or 6 exceeds the number of visible tokens) reduces GFLOPs by 12-18\% compared to $r$=0, with minimal performance loss (-0.05\% to -0.1\%). Secondly, increasing the hard mask ratio from 25\% to 70\% under the same $r$ improves classification WAR by about 1\% while reducing GFLOPs by 40-50\%. These results confirm that adaptively reducing noisy or redundant video information effectively lowers computational costs and improving performance.

\subsection{Visualization}
\begin{figure}[t]
\centerline{\includegraphics[scale=0.34]{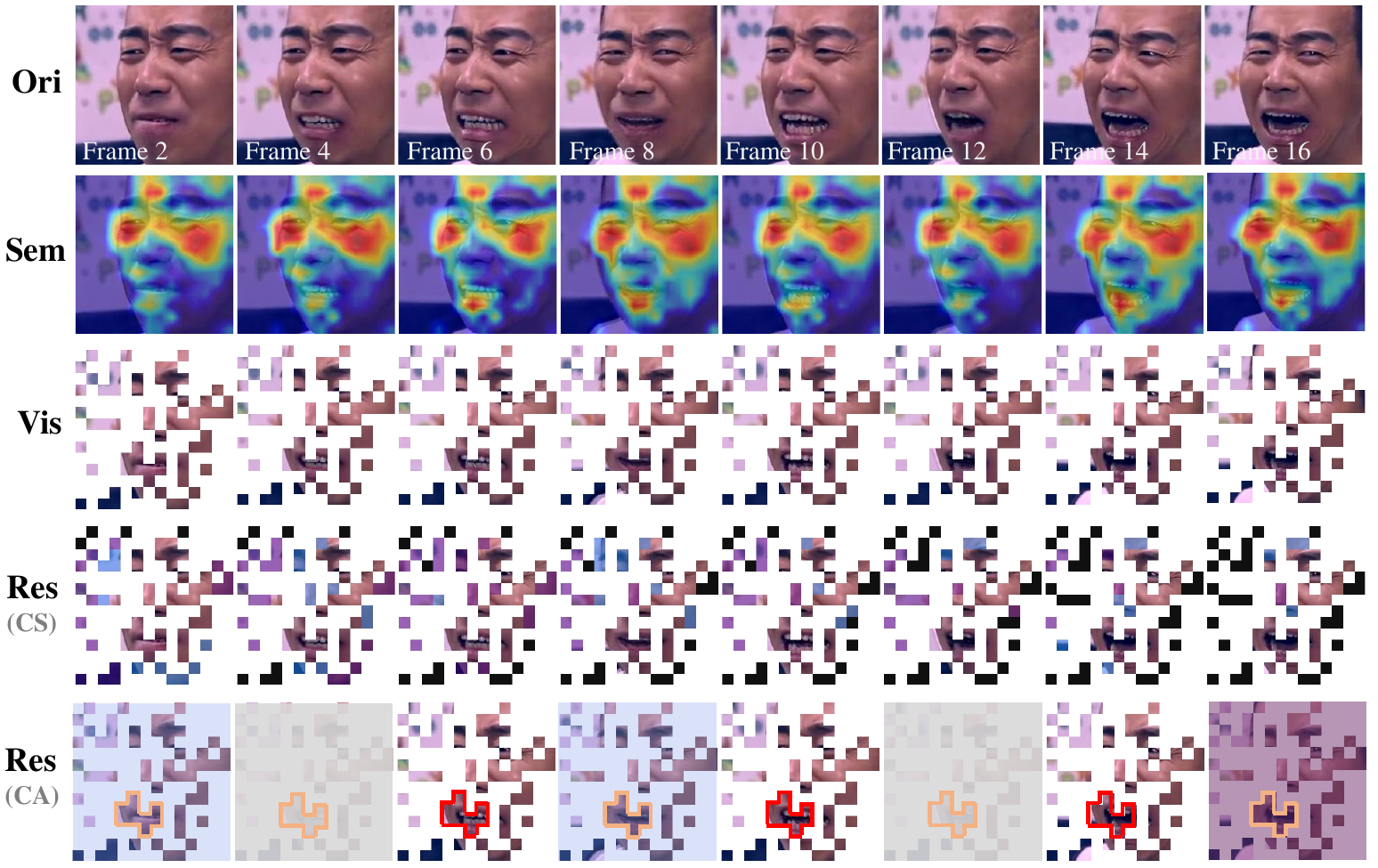}}
\caption{\textbf{Visualization of class activation maps and temporal soft masks.} 
Row 1: Original frames. Row 2: Class-semantic activation maps. Row 3: Visible patches after hard masking. Row 4: Class-semantic (CS) soft mask on visible patches. Row 5: Class-agnostic (CA) dynamic soft mask. Key dynamic regions are in \textcolor{red}{red} and \textcolor{orange}{orange}. Colors from \textcolor{Blue}{blue} to \textcolor{violet}{purple} to \textcolor{gray}{gray}/black indicate increasing mask intensity.}
\vspace{-2em}
\label{fig:vis}
\end{figure}

In Row 2 of Fig.~\ref{fig:vis}, our model effectively identifies key facial regions via class-semantic activation maps. For sadness, it consistently highlights the cheeks and eyebrows, with variable attention to the mouth. In Row 4, redundancy accumulates over time, as seen in the increasing excluded tokens (black) from frame 2 to 16. The class-semantic soft mask adapts based on both feature similarity and classification semantics (Row 2); for instance, cheek regions receive a lighter mask despite their similarity. In Row 5, the class-agnostic dynamic soft mask emphasizes key (6, 10, 14), capturing expression shifts. Red and orange boxes mark changes, confirmed in Row 1.
\section{Conclusion}
In this work, we presented \modelname, a novel supervised temporal soft masked autoencoder network for DFER task. By combining a self-supervised reconstruction branch with a supervised classification branch, \modelname{} efficiently mitigates irrelevant information and reduces redundancy through adaptive temporal soft masking. Our model improves computational efficiency without sacrificing performance, as demonstrated through extensive experiments on standard benchmarks.

\bibliographystyle{IEEEbib}
\bibliography{icme2025references}

\end{document}